\documentclass[letterpaper, 10 pt, conference]{ieeeconf}  

\IEEEoverridecommandlockouts                              

\usepackage{graphicx} 

\usepackage{xcolor}
\usepackage{algpseudocode}
\usepackage{algorithm}
\usepackage{times}

\usepackage{mathtools}
\usepackage{xcolor} 
\usepackage{ulem} 
\usepackage{multirow}
\usepackage{multicol}
\usepackage{bbm}
\usepackage{comment}

\usepackage{color}
\usepackage{latexsym}
\usepackage{multicol}

\usepackage[font=small,labelfont=bf]{caption}
\usepackage{lipsum}
\usepackage{subcaption}

\newcommand{\Next}{\bigcirc}

\newcommand{\Event}{\diamondsuit}
\newcommand{\Implies}{\Rightarrow}
\newcommand{\Then}{\mathcal{T}}
\newcommand{\True}{\top}

\newtheorem{definition}{Definition}
\newtheorem{problem}{Problem}

\newtheorem{example}{Example}
\newcommand*{\mydprime}{^{\prime\prime}\mkern-1.2mu}

\usepackage{amsmath} 
\usepackage{amssymb}  
\usepackage{dsfont}
\newcommand{\ignore}[1]{%
}

\title{Automata Guided Reinforcement Learning With Demonstrations}

\author{Xiao Li, Yao Ma and Calin Belta
  \thanks{X. Li, Y. Ma and C. Belta are with the Dept. of Mechanical Engineering,
          Boston University, Boston, MA, USA. \{xli87, yaoma, cbelta\}@bu.edu}
}

\begin{document}

\maketitle
\thispagestyle{empty}
\pagestyle{empty}

\begin{abstract}
Tasks with complex structures and long horizons pose a challenge for reinforcement learning agents due to the difficulty in specifying the task in terms of reward functions as well as large variances in the learning signal. We propose to address these problems by combining temporal logic (TL) with reinforcement learning from demonstrations. Our method automatically generates intrinsic rewards that align with the overall task goal given a TL task specification. The policy resulting from our framework has an interpretable and hierarchical structure. We validate the proposed method experimentally on a set of robotic manipulation tasks.
\end{abstract}

\section{Introduction}

Learning robotic skills for tasks with complex structures and long horizons poses a significant challenge for current reinforcement learning methods. Recent endeavors have focused mainly on lower level motor control tasks such as grasping~\cite{Mahler2017DexNet3C} \cite{Kalashnikov2018QTOptSD}, dexterous hand manipulation\cite{openaiHand}, lego insertion~\cite{Haarnoja2018ComposableDR}. However, demonstration of robotic systems capable of learning controls for tasks that require logical execution of subtasks has been less successful. 

The first challenge in learning of complex tasks is low initial success rates. The agent rarely receives a positive reward signal through exploration. It has been shown that providing demonstration data can significantly facilitate learning. This idea has been adopted to learn tasks such as block stacking~\cite{Nair2017OvercomingEI}\cite{Zhu2018ReinforcementAI}, insertion~\cite{Lin2016RobotLF}\cite{Vecerik2017LeveragingDF} as well as autonomous driving~\cite{Silver2012LearningAD}\cite{Pan2010AgileAD}. Intrinsic rewards have also been shown to provide extra learning signal~\cite{Singh2004IntrinsicallyMR}, however, carelessly designed intrinsic rewards can adversely effect learning performance.

Learning only from demonstrations suffers from covariate shift (accumulative error resulting from deviation of state and action distributions from demonstrations) which is often addressed by combining demonstrations with reinforcement learning~\cite{rajeswaranlearning}\cite{Peng2018DeepMimicED}. However, tasks that require long sequences of actions to complete usually results in high variance learning signals (gradients) which drastically hinders the learning progress. This problem can be alleviated by using temporal abstractions~\cite{Suttona1998BetweenMA}. Hierarchical reinforcement learning has recently been successfully applied in simulated control tasks~\cite{Nachum2018DataEfficientHR}, simple navigation tasks~\cite{Bischoff2013HierarchicalRL} and robotic manipulation tasks~\cite{Gudimella2017DeepRL}\cite{Xu2017NeuralTP}.

\begin{figure}[h]
\vspace{-0.1in}
\begin{center}
\includegraphics[width=0.75\linewidth]{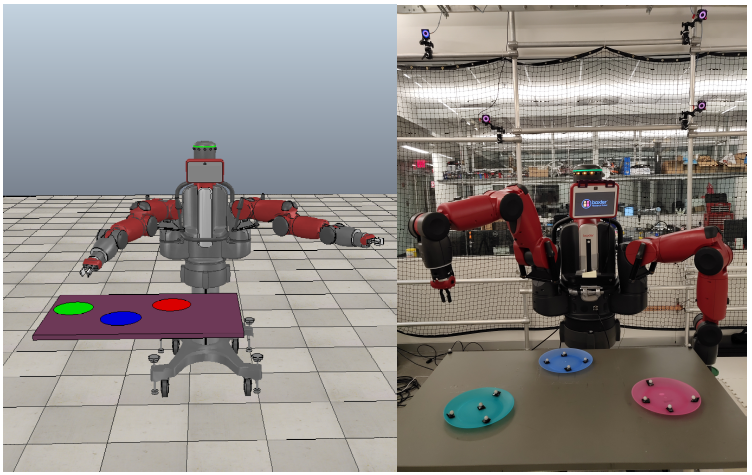}
\vspace{0.1in}
\caption{\textbf{left}: Training environment in the V-REP simulator \cite{Rohmer2013VREPAV}. \textbf{right}: Experimental environment.}\label{fig:6.1}
\end{center}
\end{figure}

The third challenge in learning complex tasks arises with the specification of task rewards. Reward engineering is time-consuming for low-level control tasks where efforts in reward shaping~\cite{Ng1999PolicyIU} and tuning are often necessary. This process is much more difficult when the structure of the tasks complicates. Authors of~\cite{Popov2017DataefficientDR} have shown that it is already a considerable effort to specify the reward function for simple block-stacking tasks. 

We propose to address the above problems by using formal specification languages, particularly temporal logic (TL) as the task specification language. TL has been used in control synthesis~\cite{andersson2017control}, path planning~\cite{leahy2017informative} and learning~\cite{Chen2012LTLRM}\cite{kasenberg2017interpretable}. It has been shown to provide convenience and performance guarantees in tasks with logical structures and persistence requirements. 

Our goal in this work is to provide a framework that integrates temporal logic with reinforcement learning from demonstrations. We show that our framework generates intrinsic rewards that are aligned with the task goals and results in a policy with interpretable hierarchy. We experimentally validate our framework on learning of robotic manipulation tasks with logical structures. All of our training is done in the simulation environment as shown in Figure~\ref{fig:6.1}. We show that when configured properly, the policies can transfer directly to the real robot.

\section{Related Work}
\label{sec:7}

Policy/task sketches have been used to decompose a complex task to a set of sub-tasks~\cite{Andreas2017ModularMR}\cite{Shiarlis2018TACOLT}. However, these methods only support sequential execution of the subtasks whereas our approach is able to compose subtasks in any temporal/logical relationships. Moreover, given the specification of the task in syntactically co-safe truncated linear temporal logic (scTLTL), our method does not require specification of each subtask in terms of reward functions.

The works in~\cite{Wen2017LearningFD}\cite{icarte2018using} are the most related to ours. Authors of~\cite{Wen2017LearningFD} incorporates maximum-likelihood inverse reinforcement learning with side information (addition constraints of the task) in the form of co-safe linear temporal logic (which is transformed to an equivalent finite state automaton). However, their methods only support discrete state and action spaces. The authors of~\cite{icarte2018using} propose the reward machine which in effect is an FSA. However, the user is required to manually design the reward machine whereas our method generates the reward machine from TL specifications. 

\section{Preliminaries}
\label{sec:2}
\subsection{Off-Policy Reinforcement Learning}
\label{subsec:2.2}

We start with the definition of a Markov Decision Process.

\begin{definition}\label{def2}
An MDP is defined as a tuple $\mathcal{M} = \langle S,A,p(\cdot|\cdot,\cdot),r(\cdot,\cdot, \cdot)\rangle$, where $S\subseteq {\rm I\!R}^n$ is the state space ; $A \subseteq {\rm I\!R}^m$ is the action space ($S$ and $A$ can also be discrete sets); $p: S \times A \times S \to [0,1]$ is the transition function with $p(s^{\prime}|s,a)$ being the conditional probability density of taking action $a \in A$ at state $s \in S$ and ending up in state $s^{\prime} \in S$; $r: S \times A \times S \to {\rm I\!R}$ is the reward function with $r(s,a,s^\prime)$ being the reward obtained by executing action $a$ at state $s$ and transitioning to $s^\prime$. 
\end{definition}

We define a task to be the process of finding the optimal policy $\pi^\star: S \to A$ (or $\pi^\star: S \times A \to [0,1]$ for stochastic policies) that maximizes the expected return, i.e.

\begin{equation}\label{eq2B1}
\pi^\star = \underset{\pi}{\arg\max}\mathbb{E}^\pi[\sum_{t=0}^{T-1} r(s_t, a_t, s_{t+1})],
\end{equation}
The horizon of a task (denoted $T$) is defined as the maximum allowable time-steps of each execution of $\pi$ and hence the maximum length of a trajectory. 
In Equation~\eqref{eq2B1}, $\mathbb{E}^\pi[\cdot]$ is the expectation following $\pi$. The state-action value function is defined as 

\begin{equation}\label{eq2B2}
Q^\pi(s,a) = \mathbb{E}^\pi[\sum_{t=0}^{T-1} r(s_t, a_t, s_{t+1})|s_0=s, a_0=a]
\end{equation}
\noindent i.e. it is the expected return of choosing action $a$ at state $s$ and following $\pi$ onwards. For off-policy actor critic methods such as deep deterministic policy gradient~\cite{Lillicrap2015ContinuousCW}, $Q^\pi$ is used to evaluate the quality of policy $\pi$. Parameterized $Q^\pi_w$ and $\pi_\theta$ ($w$ and $\theta$ are learnable parameters) are optimized alternately to obtain $\pi^\star_\theta$.

\subsection{scTLTL and Finite State Automata}
\label{subsec:2.1}
We consider tasks specified with \textit{syntactically co-safe Truncated Linear Temporal Logic} (scTLTL) which is derived from truncated linear temporal logic(TLTL)~\cite{li2017reinforcement}. The $\Box$~(always) operator is omitted in order to establish a connection between TLTL and finite state automaton (Definition~\ref{def1}).  The syntax of scTLTL is defined as 

\begin{equation}\label{eq2A1}
\begin{split}
\phi := \ & \True \,\,|\,\, f(s) < c \,\,| \,\, \neg \phi \,\,|\,\, \phi \wedge \psi \,\,|\,\, \\ & \Event \phi  \,\,|\,\, \phi \, \mathcal{U} \, \psi \,\,|\,\, \phi\, \Then\, \psi \,\,|\,\, \Next \phi \,\, 
\end{split}
\end{equation}
where $\True$ is the True Boolean constant. $s \in S$ is a MDP state in Definition~\ref{def2}.  $f(s) < c$ is a predicate over the MDP states where $c \in {\rm I\!R}$.
$\neg$~(negation/not), $\wedge$~(conjunction/and) are
Boolean connectives.
$\Event$~(eventually),  $\mathcal{U}$~(until), $\Then$~(then), $\Next$~(next),
are temporal operators.$\Implies$ (implication) and and $\vee$~(disjunction/or) can be derived from the above operators. 

We denote $s_t \in S$ to be the state at time $t$, and $s_{t:t+k}$ to be a sequence of states
(state trajectory) from time $t$ to $t+k$, i.e., $s_{t:t+k}=s_ts_{t+1}...s_{t+k}$. The Boolean semantics of scTLTL is defined as:

\begin{alignat*}{3}
&s_{t:t+k} \models f(s)<c \quad &&\Leftrightarrow \quad &&f(s_t) <c, \\
&s_{t:t+k} \models \neg \phi \quad &&\Leftrightarrow \quad &&\neg(s_{t:t+k}\models \phi),\\
&s_{t:t+k} \models \phi \Rightarrow \psi  \quad &&\Leftrightarrow \quad && (s_{t:t+k} \models \phi) \Rightarrow (s_{t:t+k} \models \psi),\\
&s_{t:t+k} \models \phi \wedge \psi \quad &&\Leftrightarrow \quad && (s_{t:t+k} \models \phi) \wedge (s_{t:t+k} \models \psi),\\
&s_{t:t+k} \models \phi \vee \psi \quad &&\Leftrightarrow \quad && (s_{t:t+k} \models \phi) \vee (s_{t:t+k} \models \psi),\\
&s_{t:t+k} \models \Next \phi  \quad &&\Leftrightarrow \quad && (s_{t+1:t+k} \models \phi) \wedge (k>0), \\
&s_{t:t+k} \models \Event \phi \quad &&\Leftrightarrow \quad && \exists t^\prime \in [t,t+k) \ s_{t^\prime:t+k} \models \phi,\\
&s_{t:t+k} \models \phi \,\, \mathcal{U} \,\, \psi \quad &&\Leftrightarrow \quad && \exists t^\prime \in [t,t+k) \,\,s.t.\,\, s_{t^\prime:t+k} \models \psi \\
&\,&&\,&& \wedge (\forall t^{\prime\prime} \in [t,t^\prime) \ s_{t^{\prime\prime}:t^\prime} \models \phi),\\
&s_{t:t+k} \models \phi \,\, \mathcal{T}\,\, \psi \quad &&\Leftrightarrow \quad && \exists t^\prime \in [t,t+k) \,\,s.t.\,\, s_{t^\prime:t+k} \models \psi \\
&\,&&\,&& \wedge (\exists t^{\prime\prime} \in [t,t^\prime) \ s_{t^{\prime\prime}:t^\prime} \models \phi).
\end{alignat*}
A trajectory $s_{0:T}$ is said to satisfy formula $\phi$ if $s_{0:T} \models \phi$.

There exists a real-valued function $\rho(s_{0:T}, \phi)$ called robustness degree (sometimes referred to as just robustness) that measures the level of satisfaction of trajectory $s_{0:T}$ with respect to a scTLTL formula $\phi$. The robustness can be defined recursively as 

\begin{alignat*}{3}
&\rho(s_{t:t+k}, \True)\quad && = \quad && \rho_{max},\\
&\rho(s_{t:t+k},f(s_t)<c) \quad && = \quad &&c-f(s_t),\\
&\rho(s_{t:t+k},\neg \phi) \quad && = \quad &&-\rho(s_{t:t+k},\phi),\\
&\rho(s_{t:t+k}, \phi \,\, \Rightarrow \psi) \quad && = \quad && \max(-\rho(s_{t:t+k}, \phi), \rho(s_{t:t+k}, \psi))\\
&\rho(s_{t:t+k},\phi_1\wedge \phi_2) \quad &&= \quad &&\min(\rho(s_{t:t+k},\phi_1),\rho(s_{t:t+k},\phi_2)), \\
&\rho(s_{t:t+k},\phi_1\vee \phi_2) \quad &&= \quad &&\max(\rho(s_{t:t+k},\phi_1),\rho(s_{t:t+k},\phi_2)),\\
&\rho(s_{t:t+k}, \Next \phi) \quad && = \quad && \rho(s_{t+1:t+k},\phi) \,\,(k>0), \\
&\rho(s_{t:t+k},\Event \phi) \quad &&= \quad && \underset{t^{\prime} \in [t,t+k)}{\max}(\rho(s_{t^{\prime}:t+k},\phi)),\\
&\rho(s_{t:t+k},\phi \,\, \mathcal{U} \,\, \psi) \quad && = \quad && \underset{t^{\prime} \in [t,t+k)}{\max}( \min (\rho(s_{t^{\prime}:t+k},\psi), \\
& \, &&\, && \underset{t^{\mydprime} \in [t,t^{\prime})}{\min}\rho(s_{t\mydprime:t^\prime},\phi))),\\
&\rho(s_{t:t+k},\phi \,\, \Then \,\, \psi) \quad && = \quad && \underset{t^{\prime} \in [t,t+k)}{\max}( \min (\rho(s_{t^{\prime}:t+k},\psi), \\
& \, &&\, && \underset{t^{\mydprime} \in [t,t^{\prime})}{\max}\rho(s_{t\mydprime:t^\prime},\phi))),
\end{alignat*}

where $\rho_{max}$ represents the maximum robustness value. A robustness of greater than zero implies that $s_{t:t+k}$ satisfies $\phi$ and vice versa ($\rho(s_{t:t+k},\phi) > 0 \Rightarrow s_{t:t+k} \models \phi$ and
$\rho(s_{t:t+k},\phi) < 0 \Rightarrow s_{t:t+k} \not\models \phi$). The robustness can substitute Boolean semantics to enforce
the specification $\phi$.


\begin{definition}\label{def1} 
An FSA corresponding to a scTLTL formula $\phi$ \footnote{Here we slightly modify the conventional definition of FSA and incorporate the probabilities in Equations~\eqref{eq2A2}. For simplicity, we continue to adopt the term FSA.} is defined as a tuple  $\mathcal{A}_\phi=\langle \mathbb{Q}_\phi, \Psi_\phi, q_{0}, p_\phi(\cdot | \cdot), \mathcal{F}_\phi \rangle $, where $\mathbb{Q}_\phi$ is a set of automaton states; $\Psi_\phi$ is the input alphabet (a set of first order logic formula); $q_{0} \in \mathbb{Q}_\phi$ is the initial state; $p_\phi:\mathbb{Q}_\phi \times \mathbb{Q}_\phi \rightarrow [0,1]$ is a conditional probability defined as 

\begin{equation}\label{eq2A2}
\begin{split}
p_\phi(q_j | q_i) &= \begin{cases}
1 & \psi_{q_i, q_j} \textrm{ is true} \\
 0 & \text{otherwise}.
 \end{cases} \\
 \quad\quad\quad\qquad or  \\
 p_\phi(q_j | q_i, s) &= \begin{cases}
1 & \rho(s,\psi_{q_i, q_j})>0\\
 0 & \text{otherwise}.
 \end{cases} 
\end{split}
\end{equation}

\noindent $\mathcal{F}_\phi$ is a set of final automaton states. The transitions in the FSA are deterministic. For reasons that will become clear later, we adopt the probability notation in Equation~\eqref{eq2A2} so that we can combine it with an MDP transition.

\end{definition}

We denote $\psi_{q_i, q_j} \in \Psi_\phi$ the predicate guarding the transition from $q_i$ to $q_j$. Because $\psi_{q_i, q_j}$ is a predicate without temporal operators, the robustness $\rho(s_{t:t+k}, \psi_{q_i, q_j})$ is only evaluated at $s_t$. Therefore, we use the shorthand $\rho(s_{t}, \psi_{q_i, q_j}) = \rho(s_{t:t+k}, \psi_{q_i, q_j})$. The translation from a TLTL formula to a FSA can be done automatically with available packages like Lomap \cite{lomap}. An example of scTLTL is provided in the next section.

\section{Problem Formulation and Approach}
 \label{sec:3}
 \label{subsec:3.1}
 
\begin{problem}\label{p1}
Given an MDP  $\mathcal{M} = \langle S,A,p(\cdot|\cdot,\cdot),r(\cdot,\cdot, \cdot)\rangle$ with unknown transition dynamics $p(\cdot|\cdot,\cdot)$ and a scTLTL formula $\phi$ as in Definition~\ref{def1}, find a policy $\pi^\star_\phi$ such that


\begin{equation}\label{eq3A1}
\pi^\star_\phi = \underset{\pi_\phi}{\arg \max}\mathbb{E}^{\pi_\phi}[\mathds{1}(\rho(s_{0:T}, \phi)>0)]. 
\end{equation}
\noindent where $\mathds{1}(\rho(s_{0:T}, \phi)>0)$  is an indicator function with value $1$ if $\rho(s_{0:T}, \phi)>0$ and $0$ otherwise.
\end{problem}

 $\pi^\star_\phi$ in Equation~\eqref{eq3A1} is said to satisfy $\phi$. Problem~\ref{p1} defines a policy search problem where the trajectories resulting from following the optimal policy should satisfy the given scTLTL formula in expectation. On a high level, our approach is to construct a product MDP between $\mathcal{M}$ and $\mathcal{A}_\phi$ and learn policy $\pi_\phi$ using the product. To accelerate learning, we provide human demonstrations of the task specified by $\phi$ and provide a simple technique to transform the demonstrations compatible with the product MDP.
 
 \section{FSA Augmented MDP}
 \label{subsec:3.2}

We introduce the FSA augmented MDP:

\begin{definition}\label{def3}
 An FSA augmented MDP corresponding to scTLTL formula $\phi$ (constructed from FSA $\langle \mathbb{Q}_\phi, \Psi_\phi, q_{0}, p_\phi(\cdot | \cdot), \mathcal{F}_\phi \rangle$ and MDP $\langle S,A,p(\cdot|\cdot,\cdot),r(\cdot,\cdot, \cdot)\rangle$) is defined as $\mathcal{M} _\phi= \langle \tilde{S}, A, \tilde{p}(\cdot|\cdot,\cdot),\tilde{r}(\cdot, \cdot),  \mathcal{F}_\phi \rangle$ where $\tilde{S} \subseteq S \times \mathbb{Q}_{\phi}$, $\tilde{p}(\tilde{s}'|\tilde{s},a)$ is the probability of transitioning to $\tilde{s}^\prime$ given $\tilde{s}$ and $a$,

 \begin{equation}\label{eq3A2}
 \tilde{p}(\tilde{s}'|\tilde{s},a) = p\big((s', q')|(s,q), a\big)
 = \begin{cases}
 p(s'|s,a) & p_\phi(q'|q,s) =1 \\
 0 & \text{otherwise}.
 \end{cases}
 \end{equation}

\noindent $p_\phi$ is defined in Equation~\eqref{eq2A2}. $\tilde{r}: \tilde{S} \times \tilde{S} \to {\rm I\!R}$ is the FSA augmented reward function, defined by 


\begin{equation}\label{eq3A3}
\tilde{r}(\tilde{s},\tilde{s}') = \rho(s',D_{\phi}^{q}),
\end{equation}
\noindent where $D_{\phi}^{q}=\bigvee_{q^\prime \in\Omega_{q}} \psi_{q, q^\prime}$ represents the disjunction of all predicates guarding the transitions that originate from $q$ ($\Omega_{q}$ is the set of automata states that are connected with $q$ through outgoing edges). Equation~\eqref{eq3A3} effectively acts as an intrinsic reward that aligns with the overall goal of Equation~\eqref{eq3A1}. 
\end{definition}

\begin{figure}[h]
\vspace{-0.1in}
\begin{center}
\includegraphics[width=0.8\linewidth]{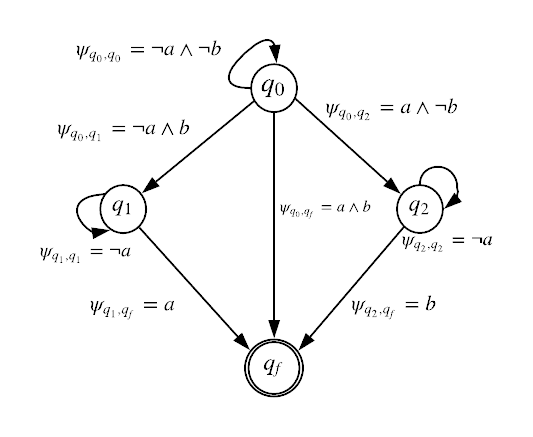}
\vspace{-0.1in}
\caption{Finite state automaton generated from formula $\diamondsuit a \wedge \diamondsuit b$}\label{fig:1}
\end{center}
\end{figure}

\begin{example}
Figure~\ref{fig:1} illustrates the FSA resulting from formula $\phi=\diamondsuit a \wedge \diamondsuit b$ (where $a: s > 3  \wedge s < 5, b: s > 8 \wedge s < 10$ are predicates over states). In English, $\phi$ entails that during a run, regions specified by $a$ and $b$ need to be visited at least once.  The FSA has four automaton states $Q_{\phi}=\{q0, q_1, q_2, q_f\}$ with $q_0$ being the input(initial) state (here $q_i$ serves to track the progress in satisfying $\phi$). The input alphabet is defined as $\Psi_\phi=\{\neg a \wedge \neg b, \neg a \wedge b, a \wedge \neg b, a \wedge b\}$. Shorthands are used in the figure, for example $a = (a \wedge b) \vee (a \wedge \neg b)$. $\Psi_\phi$ represents the power set of $\{a,b\}$, i.e. $\Psi_\phi = 2^{\{a,b\}}$.  During execution, the FSA always starts from state $q_0, s_0$ and transitions according to Equation~\eqref{eq3A2}. The specification is satisfied when $q_f$ is reached.
\end{example}

The goal is to find the optimal policy that maximizes the expected sum of discounted return, i.e.

\begin{equation}\label{eq3A4}
 \pi^\star_\phi = \underset{\pi_\phi}{\arg\max} \mathbb{E}^{\pi_\phi}\left[\sum_{t=0}^{T-1} \gamma^{t+1}\tilde{r}(\tilde{s}_t,\tilde{s}_{t+1})\right], 
\end{equation}

\noindent where $\gamma < 1$ is the discount factor, $T$ is the time horizon.

The reward function in Equation~\eqref{eq3A3} encourages the system to exit the current automaton state and move on to the next, and by doing so eventually reach the final state $q_f$ (property of FSA) which satisfies the TL specification and hence Equation~\eqref{eq3A1}. The discount factor in Equation~\eqref{eq3A4} reduces the number of satisfying policies to one. 

The FSA augmented MDP can be constructed with any standard MDP and a scTLTL formula, and Equation~\eqref{eq3A4} can be solved with any off-the-shelf RL algorithm. After obtaining the optimal policy $\pi_\phi^\star$, executing $\pi_\phi^\star(s_t,q_i)$ without transitioning the automaton state (i.e. keeping $q_i$ fixed) results in a set of meaningful policies that can be used as is or composed with other such policies.

\section{FSA Guided Reinforcement Learning From Demonstrations}
In this section, we introduce our main algorithm - FSA guided reinforcement learning from demonstrations. The algorithm takes as input a scTLTL formula $\phi$, a randomly initialized policy $\pi_\theta(s,q)$ and a set of demonstration trajectories $D = \{\tau_i\}, i \in 1,...,n$ that satisfy $\phi$, where $\tau = (s_0, a_0, ..., s_T)$ is the state-action trajectory. The algorithm consists of the following steps:

\begin{enumerate}
    \item Construct the FSA augmented MDP $\mathcal{M}_\phi$. 
    \item For each demonstration trajectory $\tau_i$, construct the Q-appended demonstration trajectory $\tau^Q_i = (s_0, a_0, q_0, ..., s_T)$ by finding the corresponding $q_t$ for each $(s_t, a_t)$ using Equation~\eqref{eq2A2}. Denote $\mathcal{D}^Q = \{\tau^Q_i\}, i = 1, .., n$.
    \item Perform behavior cloning (supervise learning on the demonstration trajectories) to initialize policy (details provided in Section~\ref{sec:algo_details}).
    \item Train the agent using any reinforcement learning from demonstration algorithm (such as \cite{Peng2018DeepMimicED}, \cite{Nair2017OvercomingEI}, \cite{rajeswaranlearning}).
\end{enumerate}

Algorithm~\ref{alg:1} shows each step with its inputs and output. We will discuss our choices of behavior cloning and RL algorithms in the next section. 

\begin{algorithm}
\caption{FSA Guided Reinforcement Learning From Demonstrations}
\label{alg:1}
\begin{algorithmic}[1]
\State \textbf{Inputs}: scTLTL task specification $\phi$, randomly initialized policy $\pi_{\theta}$, a set of $n$ demonstration trajectories $\mathcal{D} = \{\tau_i\}, i \in 1,...,n$.
\State Construct the FSA augmented MDP $\mathcal{M}_\phi$ 
\State $\mathcal{D}^Q \leftarrow \textit{ConstructQAppendedDemoBatch}(\mathcal{M}_\phi, \mathcal{D})$
\State $\theta \leftarrow \textit{BehaviorCloning}(\theta, \mathcal{D}^Q)$
\State $\theta^\star \leftarrow \textit{RLfD}(\mathcal{M}_\phi, \mathcal{D}^Q)$ \Comment{$RLfD$ stands for any learning from demonstration algorithm}
\end{algorithmic}
\end{algorithm}

\begin{figure}[h]
\vspace{-0.1in}
\begin{center}
\includegraphics[width=0.8\linewidth]{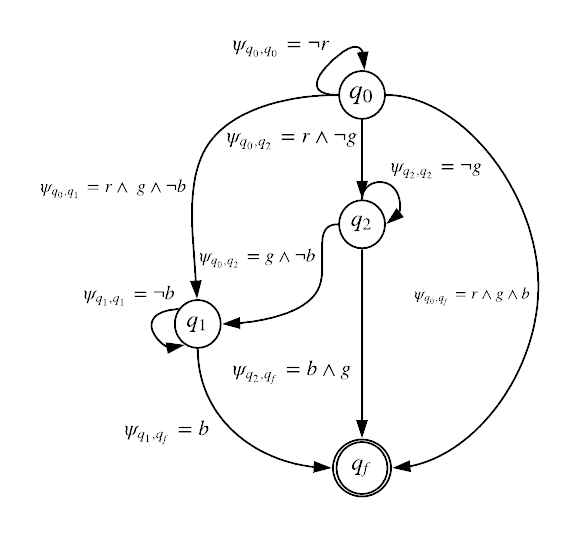}
\vspace{-0.1in}
\caption{Finite state automaton generated from formula $\diamondsuit (r \wedge \diamondsuit ( g \wedge \diamondsuit b)) $.}\label{fig:3}
\end{center}
\end{figure}


\begin{figure*}
\label{fig:3}
\centering
\begin{multicols}{2}
\includegraphics[width=2.\linewidth]{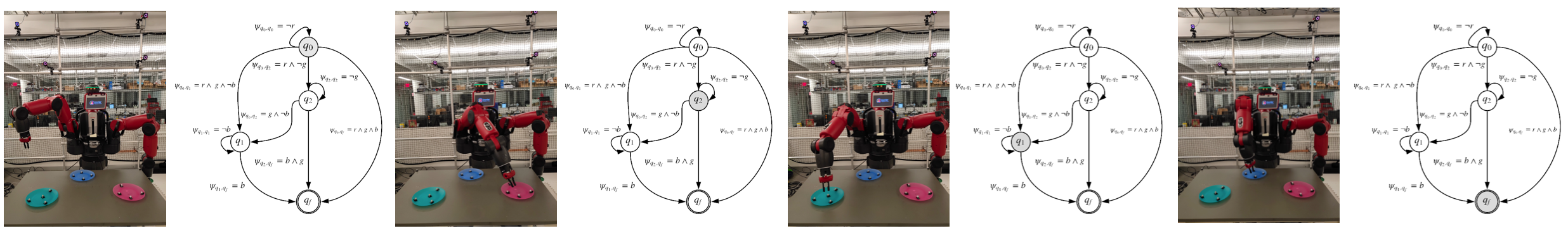}
\end{multicols}
\caption{Sample execution of task 1: $\diamondsuit (\psi^{red} \wedge \diamondsuit ( \psi^{green} \wedge \diamondsuit \psi^{blue}))$ with FSA (same as Figure 3) transitions shown. The shaded $q$ state represents the current automaton state.} \label{fig:7.1}
\end{figure*}

In this section we present some preliminary experimental results using the FSA augmented MDP to learn temporal logic specified tasks.

\section{Experiments}
\label{sec:4}

\subsection{Experiment Setup}
As shown in Figure~\ref{fig:7.1}, we control one arm of a Baxter robot (7 degrees of freedom) to traverse among three regions defined by the red, green and blue disks. The positions of the disks are tracked by our motion capture system and thus fully observable. Our state space is 16 dimensional that includes 7 joint angles and the three disk positions relative to the gripper (9 dimensional) denoted by $\boldsymbol{p}^{red}, \boldsymbol{p}^{green}, \boldsymbol{p}^{blue}$. Our action space is the 7-dimensional joint velocities. We define three predicates $\psi_i = |\boldsymbol{p}^i| < \epsilon, i \in \{red, green, blue\}$, $\epsilon$ is a threshold which we set to be 5 centimeters.

We test our algorithm on two tasks 

\begin{itemize}
    \item Task 1: $\phi_1 = \diamondsuit (\psi^{red} \wedge \diamondsuit ( \psi^{green} \wedge \diamondsuit \psi^{blue}))$ \\
    Description: visit regions red, green, and blue in this order.
    \item Task 2: $\phi_2 = \diamondsuit \psi^{red} \wedge \diamondsuit \psi^{green} \wedge \diamondsuit \psi^{blue}$ \\
    Description: Eventually visit regions red, green and blue. Order does not matter.
\end{itemize}

\noindent Figure 3 shows the FSA resulting from $\phi_1$. The FSA for $\phi_2$ is similar in nature to that presented in Figure~\ref{fig:1} and therefore not included due to space constraints.

\subsection{Algorithm Details}
 \label{sec:algo_details}
For each task, we collect 50 human demonstration state-action trajectories (each demonstration about 12 seconds long) with randomized initial conditions (arm configuration and position of the regions). Demonstrations are collected by holding Baxter's gripper in gravity compensation mode while performing the task. Behavior cloning is used to initialize the policy with the following loss function

\begin{equation}
L_{BC} = \sum_{i=0}^{N_D} ||\pi_\theta(s_i) - a_i||^2,
\end{equation}

\noindent where $\pi_\theta(s_i): S \to A$ is a deterministic policy represented by a feedforward neural network with 3 layers, each lay consisting of 100 relu units. $N_D$ is the number of samples. Other behavior cloning losses can also be used~\cite{Rusu2015PolicyD}.

We use deep deterministic policy gradient (DDPG)~\cite{Lillicrap2015ContinuousCW} as our reinforcement learning algorithm. During training, we maintain two replay buffers, one for interaction data and one for demonstration data. At each update step, we sample a batch of experience from the interaction data buffer using prioritized experience replay~\cite{Schaul2015PrioritizedER} and another batch from the demonstration data buffer and combine the two batches for one update. In addition, we modify the policy loss to be 

\begin{equation}
    L_{total} = L_{DDPG} + \lambda L_{BC},
\end{equation}

\noindent where $L_{DDPG}$ is the usual DDPG actor loss (similar technique is used in~\cite{Nair2017OvercomingEI}). During training, we linearly decay $\lambda$ from 0.8 to 0.1 over 30000 update steps to favor demonstration in the beginning and unbiased DDPG loss towards the end (similar technique is used in~\cite{rajeswaranlearning}). We set the horizon $T$ to be 100 steps (5 seconds). 5 episodes of exploration data are collected to perform 10 updates. We use a learning rate of 0.0003, a discount factor of 0.99, batch size of 32 (from both buffers).

We randomly initialize the joint angles, the automaton state as well as the positions of the regions at reset of each episode in order to achieve generalization over different configurations of the workspace. An episode resets if the gripper comes too close to the table. All of our training is performed in simulation using the V-REP platform~\cite{Rohmer2013VREPAV}. The simulation environment is calibrated to the real world workspace. We set the control frequencies in both the real and simulated robot to be 20 Hz and show that the learned policies transfer directly to the real robot without fine-tuning.

\subsection{Comparison Cases}
\label{sec:comparison}
As comparison, we introduce a binary vector $b$ with three digits. A digit in $b$ is 1 if the corresponding region has been reached at least once and 0 otherwise (i.e. $b=100$ if $\epsilon - |\boldsymbol{p}^{red}| > 0$ occurs at least once in an episode. Likewise for $b=010$ for blue and $b=001$ for green). $b$ is used to track progress towards accomplishing the task. We train each task with the following shaped reward

\begin{equation}
	r_{\phi_1} = \begin{cases}
\epsilon - |\boldsymbol p^{red} | & b = 000 \\
\epsilon - |\boldsymbol p^{green} | & b = 100 \\
\epsilon - |\boldsymbol p^{blue} | & b = 110 \\
 -2 & \text{otherwise}.
 \end{cases}
\end{equation}

\begin{equation}
	r_{\phi_2} = \begin{cases}
\max(\epsilon - |\boldsymbol p^{red}|, \epsilon - |\boldsymbol p^{green}|, \epsilon - |\boldsymbol p^{blue}|) & b = 000 \\
\max(\epsilon - |\boldsymbol p^{green} |, \epsilon - |\boldsymbol p^{blue} |) & b = 100 \\
\max(\epsilon - |\boldsymbol p^{red}|, \epsilon - |\boldsymbol p^{blue}|) & b = 010 \\
\max(\epsilon - |\boldsymbol p^{red}|, \epsilon - |\boldsymbol p^{green}|) & b = 001 \\
\epsilon - |\boldsymbol p^{blue}| & b = 110 \\
\epsilon - |\boldsymbol p^{green}| & b = 101 \\
\epsilon - |\boldsymbol p^{red}| & b = 011 \\
 -2 & \text{otherwise}.
 \end{cases}
\end{equation}

\noindent on the original MDP. We also compare cases with and without demonstration.

Due to the scale difference between rewards provided by the FSA augmented MDP and the shaped reward, we present all learning curves in terms of robustness for a clear comparison. This is because the semantics of the robustness entails that a trajectory evaluating to a higher robustness value achieves better satisfaction of the TL specification (a value greater than zero guarantees satisfaction).

We acknowledge that for any given task, a well-shaped reward that accelerates learning can be provided if enough effort goes into the design and tuning process. However, this effort grows quickly with the complexity of the task. Our goal is to use formal languages to free users of this burden while achieving similar sample efficiency as a shaped reward.  

\section{Results and Discussion}
\label{sec:5}

In this section, we present our experimental results along with discussions of their implications. Figure~\ref{fig:7.1} shows an example execution of task 1 on Baxter. The automaton serves as a progress tracking mechanism that hierarchically abstracts a temporal dependent task to a set of independent ones.

\begin{figure*}
\centering
\begin{multicols}{2}
\includegraphics[width=2.\linewidth]{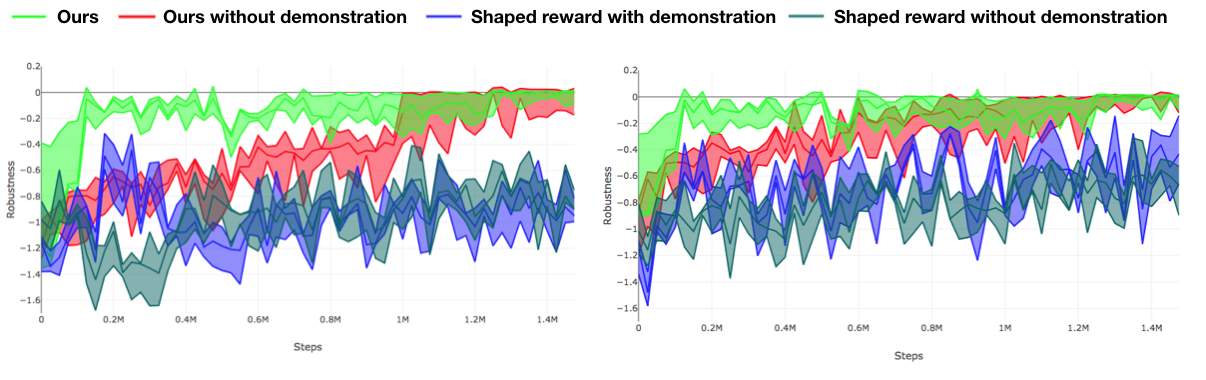}
\end{multicols}
\caption{Learning curve for \textbf{left}: Task 1 and \textbf{right}: Task 2. Steps here are referred to as environmental step}  \label{fig:8.1}
\end{figure*}

As stated in Section~\ref{sec:comparison}, since we are training with different reward functions, in order for a fair comparison, we sample a batch of 10 trajectories every 25,000 environmental steps (robot interaction step, as opposed to policy update step) and calculate the robustness for each trajectory. Their means and standard deviations are presented in Figure~\ref{fig:8.1}. In the figure, we refer to Algorithm~\ref{alg:1} as 'Ours', and learning from only FSA augmented MDP as 'Ours without demonstration'. The shaped rewards are used to train with the same learning procedure as stated in Section~\ref{sec:algo_details}. 

The results in Figure~\ref{fig:8.1} show that our method is able to solve both tasks with and without demonstrations (task is considered solved if the average robustness stabilizes above zero). However, demonstrations and behavior cloning significantly decreased the time to convergence as well as the variance during training. We can see that the agent is also able to learn very slowly using the shaped rewards but is unable to solve either task in the allocated time. The speedup of our method is mainly due to the temporal hierarchy the FSA provides. By adding one discrete dimension (the $q$ state) to the state space and randomizing on that dimension during learning, a curriculum is created to help the agent learn a set of simpler sub-tasks building up to the final task. This way the agent is able to visit various states along the task without having to first learn the correct actions leading up to those states.

\begin{figure}[h]
\vspace{-0.1in}
\begin{center}
\includegraphics[width=1.\linewidth]{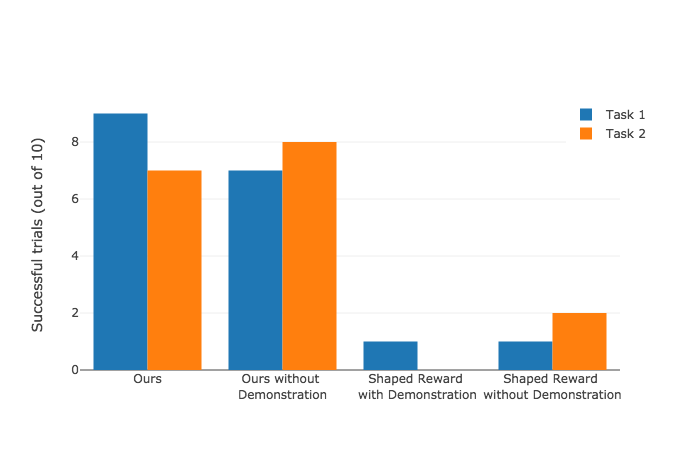}
\vspace{-0.1in}
\caption{Task success rate of the trained policies.}\label{fig:8.2}
\end{center}
\end{figure}

In all comparison cases, learning task 2 if faster than task 1. This is because task 1 imposes more constraints on the desired behavior (ordering). It is expected that even with the shaped reward, demonstration and behavior cloning is able help bootstrap learning at the initial stages. However, such initialization can be damaged as shown in Figure~\ref{fig:8.1} left. After training, we evaluate the policies by running 10 trials with randomly initialized robot and workspace configurations. Results in Figure~\ref{fig:8.2} show that the resulting policies from our method (with and without demonstrations) is able to accomplish the tasks relatively reliably whereas the policies from the shaped rewards struggled.

\begin{table}[]
\centering
\caption{Average number of steps to finish the task}
\label{table:1}
\begin{tabular}{|c|c|c|}
\hline
\textbf{}                           & \multicolumn{1}{l|}{Task 1} & \multicolumn{1}{l|}{Task 2} \\ \hline
Ours                                & 36.5                        & 34.3                        \\ \hline
Ours without demostration           & 35.7                        & 34.1                        \\ \hline
Shaped reward with demonstration    & 100                         & 93.2                        \\ \hline
Shaped reward without demonstration & 99.6                        & 95.5                        \\ \hline
\end{tabular}%
\end{table}

It should be noted that there typically will be more than one policy that satisfies Equation~\eqref{eq3A1}. However, the discount factor in Equation~\eqref{eq3A4} reduces the number of optimal policies to one (the one that yields a satisfying trajectory in the least number of steps). Table~\ref{table:1} shows the average number of steps each policy takes to accomplish the corresponding task. 


As with any formal method based technique, there is a learning curve to understanding formal languages and using them well in writing specifications. We find that the FSA has significantly helped us in understanding what we are specifying to the agent which served as an effective means to alleviate reward hacking~\cite{Amodei2016ConcretePI}. At its current state, our framework does not support specification of persistent tasks~\cite{Smith2011OptimalPP}. We have also yet to demonstrate tasks specified over MDP states and actions (e.g. if some state occurs then do something). These are possible extensions of future work. 

\section{Conclusions}
\label{sec:6}

Learning to follow logical instruction can be useful in real life (e.g. following a recipe or the traffic rules). In this work, we proposed a method to combine temporal logic with reinforcement learning from demonstrations which provides the agent with temporal hierarchy and task aligned intrinsic rewards. We showed that comparing to heuristically designed reward functions, our method provides a formalism for task specification and is able to learn with less experience. By toggling the automaton state $q$, our learned policy is able to exhibit different behaviors specified by the intrinsic reward in Equation~\eqref{eq3A3} even though no hierarchy is imposed on the policy architecture (simple feedforward neural network). For future work, we will take advantage of this characteristic and develop a set of techniques for skill composition and task-space transfer. We will also demonstrate our methods on more complex tasks.



\bibliographystyle{IEEEtran}
\bibliography{root}

\end{document}